\documentclass[preprint,12pt]{elsarticle}

\usepackage{amssymb}
\usepackage{amsmath}
\usepackage{amsthm}
\usepackage{graphicx}

\usepackage{float}

\usepackage{booktabs,siunitx}
\usepackage{multirow}

\usepackage{fullpage}

\usepackage{algpseudocode, algorithm, algorithmicx}
\newcommand*\Let[2]{\State #1 $\gets$ #2}

\biboptions{square,numbers,sort&compress}

\global\long\def\e{\mathbf{E}}
\global\long\def\one{\mathtt{1}}
\global\long\def\half{\frac{1}{2}}
\global\long\def\norm#1{\left\Vert #1\right\Vert }
\global\long\def\given#1{\left|#1\right.}

\begin{document}

\begin{frontmatter}

\title{A Derivative-Free Method for Solving Elliptic Partial Differential Equations with Deep Neural Networks}
\author{Jihun Han}
\author{Mihai Nica}
\author{Adam R Stinchcombe\corref{cor1}}
\ead{stinch@math.toronto.edu}
\cortext[cor1]{Corresponding author.}

\address{Department of Mathematics, University of Toronto, \\40 St.~George Street, Toronto ON,  M5S~2E4 Canada}

\begin{abstract}
We introduce a deep neural network based method for solving a class of elliptic partial differential equations.  We approximate the solution of the PDE with a deep neural network which is trained under the guidance of a probabilistic representation of the PDE in the spirit of the Feynman-Kac formula. The solution is given by an expectation of a martingale process driven by a Brownian motion. As Brownian walkers explore the domain, the deep neural network is iteratively trained using a form of reinforcement learning. Our method is a ``Derivative-Free Loss Method'' since it does not require the explicit calculation of the derivatives of the neural network with respect to the input neurons in order to compute the training loss. The advantages of our method are showcased in a series of test problems: a corner singularity problem, an interface problem, and an application to a chemotaxis population model.
\end{abstract}


\journal{the arXiv}

\end{frontmatter}

\section{Introduction}
With the growth of computing power and the availability of big data, machine learning, especially deep learning, has had success in a wide range of research fields such as image recognition, natural language process, and recommendation systems. Primarily, these successes are due to neural networks being readily trainable universal function approximators~\cite{UNIV}. For this same reason, neural networks can be used to describe complex physical phenomena modeled by differential equations.

Many authors have recently proposed deep learning methods to solve differential equations~\cite{ANNODEPDE,DLBSDE,HDIMPDE,DGM,PINN,SMFREE,raissi2018forward, raissi2018hidden,raissi2018hiddenNS,raissi2018deep,DEEPRITZ,ZHUUQ,DEEPBOTH} arising in various fields including fluid dynamics and quantitative finance. These methods approximate the solution of a differential equation by a deep neural network, but differ in their learning methodology and particular choice of objective or loss function. An objective function measures how well a neural network approximation satisfies the differential equation and is used as a compass to train the neural network.

In the prior work, the objective function is selected either directly from the differential equation or an equivalent formulation. This is analogous to a finite difference method (FDM) directly discretizing the differential equation and a finite element method (FEM) using the variational formulation. Sirignano and Spiliopoulos~\cite{DGM} and Raissi et al.~\cite{PINN} select the objective function so that the neural network satisfies the PDE and the boundary conditions at points within the domain. This requires the computation of the derivatives of the neural network that appear in the differential equation. Zhu et al.~\cite{ZHUUQ} includes in the loss function either a residual of the PDE or an energy functional if it is available. Similarly, Karumuri et al.~\cite{SMFREE} and Weinan and Yu~\cite{DEEPRITZ} first recast their elliptic PDEs in an equivalent variational formulation and then train a neural network to minimize the energy functional of the PDE. Sirignano and Spiliopoulos~\cite{DGM} use an efficient Monte Carlo method for computing the second-derivatives of a neural network. Weinan, Han, and Jentzen~\cite{HDIMPDE,DLBSDE} solve a class of parabolic PDEs by reformulating them as backward stochastic differential equations (BSDE). Their method is specialized to compute the solution at a single point and uses a neural network catered to the time discretization of the BSDE that is trained to satisfy the terminal conditions.

In this work, we propose a numerical method to solve a class of quasilinear elliptic PDEs. We reformulate the PDE to a probabilistic representation in the spirit of the Feynman-Kac formula, which we use to train a neural network representing the solution. In particular, the solution is given by an expectation of a martingale process driven by a Brownian motion. Unlike the Feynman-Kac formula~\cite{HUYEN}, which is an expectation involving the trajectory of a stochastic process up until its first exit time, our formulation uses an expectation over a non-random increment of time, which is a common characteristic of reinforcement learning methods. In fact, our reformulation is the Bellman equation of a Markov reward process~\cite{SUTTON} and our learning methodology is similar to deep Q-learning~\cite{DEEPQ}.  As the Brownian motions explore the domain, a neural network is iteratively trained to satisfy the Bellman equation at the positions of these walkers on every deterministic time step. This distinguishes our approach from previous probabilistic approaches based on the Feynman-Kac formula~\cite{MC1, MC2, MC3, MC4, MC5}. Moreover, our method is different from previous deep learning methods in that it is not necessary to explicitly compute the \textit{derivatives} of a neural network with respect to input neurons in solving a \textit{differential} equation.

After detailing our method in Sections~\ref{sec:prelim} and \ref{sec:Numerical_Method}, we will demonstrate its strengths. Our method is especially effective at solving problems in which the solution has singularities in its derivatives, which is shown in Section~\ref{subsec:Laplace}). Furthermore, the use of Brownian motion in our method allows us to elegantly handle jump conditions on internal boundaries, as can be seen in Section~\ref{subsec:interface}). A strength of a neural network representation is that it can represent multiple functions on multiple domains or parametrized functions by simply modifying the input layer. This is exhibited in Sections~\ref{subsec:interface} and \ref{subsec:taxis}). These examples provide convincing evidence that our method is robust, versatile, and efficient.

\section{Preliminaries}
\label{sec:prelim}
In this section, we present some background from the theory of stochastic processes and the core ideas underlying our numerical method. The details of the method and its implementation appear in Sections~\ref{sec:Numerical_Method}~and~\ref{sec:Examples}.

\subsection{Martingales and quasilinear elliptic PDEs}
There is an equivalence between the solutions of quasilinear elliptic PDEs and stochastic processes with the martingale property, which we briefly review here. We are interested in the quasilinear elliptic PDE,
\begin{align}
\mathcal{Q}(u)&:=\half\Delta u+F\cdot\nabla u-G=0 \textrm{~in~} \Omega, \label{eq:quasilinear_elliptic}
\end{align}
in which $F=F(x,u(x),\nabla u(x),\ldots) \in \mathbb{R}^d$ and $G=G(x,u(x),\nabla u(x),\ldots) \in \mathbb{R}$ may depend locally on the unknown function $u$. For ease of notation we will write $F(x,u(x))$, $G(x,u(x))$ and suppress the possible dependence on higher derivatives.
Define the stochastic process $X_t \in \mathbb{R}^d$ as a solution to the stochastic differential equation
\begin{equation}
\label{eq:stochasticprocess}
    \text{d} X_t = F(X_t, u(X_t)) \text{d} t + \text{d} B_t,
\end{equation}   
in which $B_t \in \mathbb{R}^d$ is an ordinary Brownian motion on $\mathbb{R}^d$, $F$ is the function from the PDE, Eq.~\eqref{eq:quasilinear_elliptic}, and $u:\Omega \to \mathbb{R}$ is arbitrary.

Define the stochastic process $f(t,X_t) \in \mathbb{R}$ by
\begin{equation}\label{eq:quasilinear_Martingale}
    f(t,X_t) := u(X_t) - \int_0^t G(X_s,u(X_s))\text{d} s,
\end{equation}
for arbitrary $u:\Omega \to \mathbb{R}$. Note that $f(t,X_t)$ actually depends on the entire history $\{X_s\}_{s\leq t}$ but we write it $f(t,X_t)$ for convenience of notation.

With these definitions, the following two statements are equivalent:
\begin{itemize}
    \item $u:\Omega \to \mathbb{R}$ is a solution to the PDE in Eq.~\eqref{eq:quasilinear_elliptic}: $\mathcal{Q}(u)(x) = 0$ for all $x\in \Omega$;
    \item the stochastic process $f(t,X_t)$ satisfies the martingale property 
    \begin{equation}\label{eq:martingalecondition}
        \e\left[ f(t,X_t) \given X_0 = x\right] - f(0,x) = 0 \textrm{~for all~} x\in\Omega, t>0.
    \end{equation}
\end{itemize}

 This equivalence is a standard application of It\^{o}'s lemma (see e.g. \cite{KS}), Moreover, It\^{o}'s lemma shows that the infinitesimal drift of the stochastic process $f(t,X_t)$ is precisely $\mathcal{Q}(u)$ in the sense that
 \begin{align}
    & \e\left[{f(\Delta t, X_{\Delta t})|X_0 = x}\right] - f(0,x) \nonumber\\
    & = \left( \half \Delta u(x)  +  F(x,u(x)) \cdot \nabla u(x) - G(x,u(x)) \right) \Delta t + O(\Delta t^{3/2}), \label{eq:resid}
\end{align} which holds as $\Delta t \to 0$. This drift is identically zero on $\Omega$ precisely when $u$ satisfies the PDE. Having zero drift everywhere in the domain is equivalent to $f(t,X_t)$ satisfying the martingale property. Both Eqs.~\eqref{eq:martingalecondition}~and~\eqref{eq:resid} exchange solving a PDE $\mathcal{Q}(u)=0$ into finding a martingale $f$.

In order to impose a boundary condition, for example of the form
\begin{equation}
\label{eq:BCDirichlet}
u(x) = h(x) \textrm{~on~}  \partial \Omega, 
\end{equation}
we required that $f(0,x)=h(x)$ for $x\in\partial \Omega$. This follows from Eq.~\eqref{eq:quasilinear_Martingale}, since $f(0,x)=u(x)$ for all $x$. In contrast, homogeneous Neumann conditions of the form 
\begin{equation}
\frac{\partial}{\partial n}u(x) = 0 \textrm{~on~}  \partial \Omega, 
\end{equation} 
can be imposed by reflecting $X_t$ in the normal direction on the boundary $\partial \Omega$. Mixtures between these two boundary conditions, with Dirichlet on part of $\partial \Omega$ and Neumann on another part of $\partial \Omega$, requires a straightforward modification.

\subsection{Main idea of the numerical method}
In the previous section, we transformed the search for a function $u$ that satisfies the PDE Eq.~\eqref{eq:quasilinear_elliptic} into a search for the function $f$ that has the martingale property, Eq.~\eqref{eq:martingalecondition}. As a motivating case, consider the linear elliptic boundary value problem,
\begin{equation}\label{eq:linear_elliptic}
\begin{split}
\mathcal{S}(u)&:=\sum \limits_{i,j=1}^{n}a_{ij}(x)\frac{\partial^2u}{\partial x_i\partial x_j}+\sum \limits_{i=1}^{n}b_i(x)\frac{\partial u}{\partial x_i}=g(x) \textrm{~in~} \Omega, \\
u(x) &= h(x) \textrm{~on~}  \partial \Omega. \\
\end{split}
\end{equation}
Using the equivalence from the previous section,  $u$ is the solution of Eq.~\eqref{eq:linear_elliptic} if and only if 
the stochastic process $f(t,X_t)$ is a martingale satisfying $f(0,x)=h(x)$, where $f(t,X_t)$ is defined as 
\begin{equation}
f(t,X_t):= u(X_t) - \int^{t}_{0}g(X_s)\text{d} s,
\end{equation}
and $X_t$ is the stochastic process which satisfies the stochastic differential equation, 
\begin{equation}\label{eq:linear_Martingale_process}
\text{d} X_t = \beta(X_t)\text{d} t+\alpha(X_t)\text{d} B_t, \textrm{~where~}\beta=[b_i], \textrm{~and~} \frac{1}{2}\alpha\alpha^{T}=[a_{ij}].
\end{equation}
In particular, by using the martingale property for $f$, the solution $u$ satisfies the following relation for any $t >0$ and any $X_0 \in \Omega$,
\begin{equation}\label{eq:linear_Bellman_general}
u(X_{0})=\e\left[u(X_t)-\int_{0}^{t}g(X_s)\text{d} s \right].
\end{equation}
Note that Eq.~\eqref{eq:linear_Bellman_general} is also true when $t$ is a \emph{stopping time} by application of the Doob's optional stopping time theorem (see e.g. \cite{KS}). By using the stopping time $t =  \inf\{s : X_s \not \in \Omega \}$, one obtains the well known Feynman-Kac formula for this PDE, the basis of many Monte Carlo algorithms.

Equation~\eqref{eq:linear_Bellman_general} also appears in the context of \emph{Markov reward processes} where it is known as the \emph{Bellman equation} and the unknown function $u$ is known as the \emph{value function}. Reinforcement learning \cite{SUTTON} is a subfield of machine learning that has developed very powerful tools for numerically computing the value function in situations of this form. Algorithms such as temporal difference learning \cite{SUTTON} or the deep Q-learning method \cite{DEEPQ} have been developed that apply very generally to this situation. Reinforcement learning techniques are usually applied to the even more complicated \emph{Markov decision process}, where not only is the value function unknown, but the optimal action is also unknown. Our situation is simpler because there is no unknown action policy to be learned.

In our numerical method, we will consider a class of parametrized functions $u_{\theta}(x)=u(x;\theta)$, and search for parameters $\theta$ such that Eq.~\eqref{eq:linear_Bellman_general} holds. We choose to use neural networks as our class of parametrized functions $u_\theta$, as will be described in Section~\ref{sec:Numerical_Method}. Our method is not limited to this choice - any collection of arbitrary function approximators may be used.

Our search method to find the parameters $\theta$ is inspired by techniques from reinforcement learning. We fix a timestep $\Delta t$, and construct an objective function which measures how well Eq.~\eqref{eq:linear_Bellman_general} is satisfied as a function of parameters $\theta$,
\begin{equation}\label{eq:Loss_n_linear}
\mathcal{L}(\theta) = \e_x\left[\left( u \left(x;\theta\right) - \e_{X_{0\leq t \leq \Delta t}}\left[u \left(X_{\Delta t};\theta\right) - \int_0^{\Delta t}g(X_s)\text{d} s~\Big|X_0=x \right] \right)^2\right].
\end{equation}
In Eq.~\eqref{eq:Loss_n_linear}, the inner expectation is over the trajectory and the outer expectation is over the initial locations $x$. Essentially, our algorithm performs gradient descent on this function to obtain a sequence of parameters $\theta_1,\theta_2,\ldots$ that successively reduces $\mathcal{L}(\theta)$ making $u$ closer to solving Eq.~\eqref{eq:linear_elliptic}. Section \ref{subsec:algorithm_explained} provides the exact details on how this gradient descent is implemented in our method.

Since $\Delta t$ is small, one can think of the objective function Eq.~\eqref{eq:Loss_n_linear} as approximating the value of the PDE $\mathcal{Q}(u)$ as described in Eq.~\eqref{eq:resid}. We emphasize that many other more involved methods from reinforcement learning can be applied to solve the Bellman equation Eq.~\eqref{eq:linear_Bellman_general}; the simple $L_2$ objective function we use here has this added interpretation of directly approximating the PDE. 

\subsection{The Cameron-Martin-Girsanov theorem}
The stochastic process $X_t$ explores the domain and contributes to learning the solution of the PDE at its location. The stochastic processes defined above in Eq.~\eqref{eq:stochasticprocess} or Eq.~\eqref{eq:linear_Martingale_process} might have a nontrivial drift that causes the stochastic process to be driven away from an important area of the domain. It may also have a small volatility that causes slow movement and therefore slow learning or a data imbalance in sampling and therefore an inaccurate solution. Both of these issues are discussed with respect to the example in Section~\ref{subsec:interface}.

To avoid these issues, we use an alternative martingale process $f(t,X_t)$ with $X_t$ an ordinary Brownian motion, denoted $B_t$. Being able to transformation from a possibly complicated process $X_t$ to a Brownian motion $B_t$ is a result of the Cameron-Martin-Girsanov theorem (see e.g. \cite{KS}). Applying this theorem, we find that the function $u : \Omega \to \mathbb{R}$ satisfies the PDE Eq.~\eqref{eq:quasilinear_elliptic} if and only if the stochastic process 
\begin{equation}
\begin{split}\label{eq:Cameron_Martin}
f(t,B_t) := & ~u(B_{t})\exp\Bigg(\int_{0}^{t}F(B_{s},u(B_{s}))\cdot\text{d} B_{s}-\half\int_{0}^{t}\norm{F(B_{s},u(B_{s}))}^{2}\text{d} s\Bigg)\\ &-\int_{0}^{t}G(B_{s},u(B_{s}))\text{d} s
\end{split}
\end{equation} 
is a martingale. In Eq.~\eqref{eq:Cameron_Martin}, $B_{t}\in\Omega \subset \mathbb{R}^{d}$ is a Brownian motion on the time interval $t \in [0,\infty)$ starting from an arbitrary initial condition $B_{0} \in \Omega$.  In order to remove the drift from the stochastic process $X_t$, the martingale process Eq.~\eqref{eq:Cameron_Martin} now contains an additional exponential factor, which can be thought of as discounting rewards obtained when $B_t$ happens to move in the upwind direction of the drift in $X_t$. Using the Cameron-Martin-Girsanov theorem allows the Brownian walkers to explore the domain unimpeded regardless of the PDE being solved.

From the martingale property for $f$ in Eq.~\eqref{eq:Cameron_Martin}, the solution $u$ of the Eq.~\eqref{eq:quasilinear_elliptic} satisfies the following relation for any $0<t$,
\begin{equation}\label{eq:CM_Bellman}
u(B_0)=\e[f(t, B_{t})].
\end{equation}
We search for the parametrized function $u(x;\theta)$ to minimize the residual in Eq.~\eqref{eq:CM_Bellman} by minimizing the following objective function
\begin{equation}\label{eq:loss_n_quasilinear}
\mathcal{L}(\theta) = \e_x\left[\left( u \left(x;\theta\right) - \e_{B_{0\leq t \leq \Delta t}}\left[ f_{\theta}(\Delta t, B_{\Delta t})~\Big|B_0=x \right] \right)^2\right],
\end{equation}
where $f_{\theta}$ is given by as Eq.~\eqref{eq:Cameron_Martin} with the parameterized function $u(x;\theta)$. 

\section{The numerical method}\label{sec:Numerical_Method}
This section provides the numerical algorithm to solve Eq.~\eqref{eq:quasilinear_elliptic} based on the theory described in the previous section. In particular, we choose to use a neural network as the parameterized approximation to the solution, which is trained so that Markov reward process has a value function that satisfies the Bellman equation. Note that any parametrized family of functions which are dense in the target function space could be used. This flexibility has many advantages, for example, prior knowledge of the solution (e.g. symmetry, homogeneity, etc.) can be incorporated into the choice of the parametrization. We opt to use neural networks for their ubiquity and the ease with which their parameters can be determined using gradient descent and backpropagation.

\subsection{Neural networks}\label{subsec:NeuralNetwork}
Neural networks are widely used in many applications including computer vision~\cite{BISHOPPATTERN}, natural language processing~\cite{NATURALNEURAL}, forecasting, and speech recognition~\cite{NEURALSPEECH}. The structures of neural networks are designed to extract the valuable features from data. For instance, a convolutional neural network (CNN) is suitable for image data while a recurrent neural network (RNN) is well-suited for capturing sequential information from data.

In this work, we consider the multilayer perceptron (MLP)~\cite{GOODFELLOW} and a simple variant of the residual neural network (ResNet)~\cite{RESNET} as the parameterized approximation $u(x;\theta)$ of the solution. Other architectures are compatible with our method but not explored here. In particular, the accuracy of the neural network could be improved by including problem-specific information, but we do not focus on network architecture engineering in this study.

\subsubsection{Multilayer perceptron}
Multilayer perceptron (MLP) approximator $u(x;\theta)$ with dimension $[L_0, L_1, \cdots, L_D]$ is recursively defined as 
\begin{equation}
\begin{split}\label{*}
h^{(0)} &= x, \\
h^{(i+1)}&=\sigma^{(i)}\left(W^{(i)}h^{(i)}+b^{(i)} \right),~0 \leq i \leq D-2, \\
u(x;\theta) & = W^{(D-1)}h^{(D-1)}+b^{(D-1)},
\end{split}
\end{equation}
where $W^{(i)} \in \mathbb{R}^{L_{i+1}\times L_{i}}$ is a weight matrix, $b^{(i)} \in \mathbb{R}^{L_{i+1}}$ is a bias vector, $\sigma^{(i)}$ is an activation function, and $\theta$ denotes all the weight matrices and bias vectors. The first dimension $L_0$ is generally the same as the dimension of domain $\Omega$, $d$. However, different choices of $L_0$ are made in Sections~\ref{subsec:interface} and \ref{subsec:taxis}. The last dimension $L_D$ is typically equal to $1$ since we consider scalar-valued PDEs.

\subsubsection{A variant of the residual neural network}
The residual neural network (ResNet) is widely used as a base network for feature extraction in image related applications, such as classification, object detection, and segmentation. The key of ResNet is, instead of fitting the desired mapping $\mathcal{H}(x)$ directly, first fit the residual $\mathcal{F}(x):=\mathcal{H}(x)-x$ and then recover the original $\mathcal{H}(x)$ as $\mathcal{F}(x)+x$. It is easy to implement by adding the identity shortcut connections between stacked layers and each connected block is called a residual building block. It has been empirically observed that ResNet preforms well in many applications and many researchers are interested in understanding why~\cite{THEORESNET1,THEORESNET2,THEORESNET3}. The variant of ResNet we use in this work is simply adding the identity shortcuts in a multilayer perceptron network, which is similar to the networks used elsewhere~\cite{DEEPRITZ,SMFREE,DEEPBOTH}. The following are stacks of $M$ identical residual blocks with dimension $[L_0, L_1, \cdots, L_D]$;
\begin{equation}
\begin{split}\label{eq:resnet_structure}
R^{(0)} &= \sigma^{\text{in}}\left(W^{\text{in}}x + b^{\text{in}} \right),\\
R^{(i+1)} & = \mathbf{R}(R^{(i)}),~0\leq i \leq M-1, \\
u(x;\theta) & = W^{\text{out}}R^{(M)} + b^{\text{out}}, 
\end{split}
\end{equation}
in which $\mathbf{R}$ is a residual block defined as 
\begin{equation}
\begin{split}\label{eq:resnet_block}
R^{(i,0)} &= R^{(i)}, \\
R^{(i,j+1)} &= \sigma^{(i,j)}\left(W^{(i,j)}R^{(i,j)}+b^{(i,j)}\right),~0\leq j \leq D-1, \\
\mathbf{R}\left(R^{(i)}\right) &= R^{(i)} + R^{(i,D)},
\end{split}
\end{equation}
in which $W^{\text{in}} \in \mathbb{R}^{L_{0}\times L_{\text{in}}}$,   $W^{\text{out}} \in \mathbb{R}^{L_{\text{out}}\times L_{D} }$, $W^{(i,j)} \in \mathbb{R}^{L_{j+1}\times L_{j}}$ are weight matrices, $b^{\text{in}}\in \mathbb{R}^{L_{0}}$, $b^{\text{out}}\in \mathbb{R}^{L_{\text{out}}}$ , $b^{(i,j)}\in \mathbb{R}^{L_{j+1}}$ are bias vectors, and all $\sigma$ with superscript are activation functions.  Here $L_{\text{in}}$ and $L_{\text{out}}$ are the dimension of the input and output layer and we simply set $L_0=L_D$ to avoid the linear transform between successive residual blocks.

\subsection{The algorithm}\label{subsec:algorithm_explained}

The core of the method is computing and minimizing, with gradient descent, the residual of the Bellman equation for the stochastic process $f(t,B_t)$ in Eq.~\eqref{eq:Cameron_Martin} (or Eq.~\eqref{eq:quasilinear_Martingale} if the Cameron-Martin-Girsanov theorem is not being used). The psuedocode for our method is shown in Algorithm~\ref{algo:thealgo}. Evaluating  Eq.~\eqref{eq:Cameron_Martin} with $t=\Delta t$, we see that a function $u$ is a solution if and only if $u$ satisfies
\begin{equation} \label{eq:Bellman_quasilinear}
u(B_0) = \e\left[u(B_{\Delta t})\mathcal{D}(F,u)-\mathcal{R}(G,u) \right],
\end{equation} 
in which $\mathcal{D}$ and $\mathcal{R}$ are discounts and rewards defined as 
\begin{align}\
\mathcal{D}(F,u) &= \exp\Bigg(\int_{0}^{\Delta t}F(B_{s},u(B_{s}))\cdot\text{d} B_{s}-\half\int_{0}^{\Delta t}\norm{F(B_{s},u(B_{s}))}^{2}\text{d} s\Bigg),\\
\mathcal{R}(G,u) &= \int_{0}^{\Delta t}G(B_{s},u(B_{s}))\text{d} s.
\end{align}

The goal is to find parameters $\theta$ such that the neural network $u_\theta(x) := u(x;\theta)$ satisfies Eq.~\eqref{eq:Bellman_quasilinear}. To achieve this, the training methodology is to create a sequence of parameters $\theta_1,\theta_2,\ldots $, where each set of parameters $\theta_n$ is determined from $\theta_{n-1}$ by a stochastic gradient descent update of a loss function $\mathcal{L}_n(\theta)$. The loss function $\mathcal{L}_n$ combines information from the interior of $\Omega$ obtained from a random sample of the expression in Eq.~\eqref{eq:loss_n_quasilinear} and a contribution from the boundary $\partial \Omega$.

We simulate discrete Brownian motions $B_t^{(i)}$, $i=1,\cdots,N$, which run in the interior of $\Omega$ independently and simultaneously. The interior loss function is defined as
\begin{equation}\label{eq:loss_inner}
\mathcal{L}^{\Omega}_n(\theta) := \frac{1}{N}\sum \limits_{i=1}^{N}\left(u\left(B^{(i)}_{n\Delta t};\theta\right)-y_i \right)^{2}.
\end{equation} 
The target values are
\begin{equation}\label{eq:target_value}
y_i=\hat{\e}\left[u\left(B^{(i)}_{(n+1)\Delta t};\theta_{n-1}\right)D(F,u(\cdot;\theta_{n-1}))-R(G,u(\cdot;\theta_{n-1})) ~\Big|B^{(i)}_{n\Delta t} \right],
\end{equation}
in which $D(F,u)$ and $R(G,u)$ are the discretizations of $\mathcal{D}(F,u)$ and $\mathcal{R}(G,u)$ respectively, given as 
\begin{align}
D(F,u) &= \exp \left(F(B_{0},u(B_0))\cdot (B_{\Delta t}-B_{0})-\frac{1}{2}\left\|F(B_{0},u(B_0)) \right\|^2\Delta t \right),\label{eq:Discount_Appr}\\
R(G,u) &= G(B_0,u(B_0))\Delta t.\label{eq:Reward_Appr}
\end{align}
It is important to note that the discretizations $D$ and $R$ above result in a discrete-time stochastic process $u(B_{n\Delta t})D-R$ that has the martingale property like the continuous-time process $u(B_{t})\mathcal{D}-\mathcal{R}$. Also note that the target values in Eq.~\eqref{eq:target_value} depend on the previous parameters $\theta_{n-1}$ and not the argument $\theta$ to the loss function.

It is possible that a Brownian walker will exit the domain between time 0 and $\Delta t$. We detect this by $B_{\Delta t} \not\in \Omega$, which is an admittedly biased sampling of the exit since an exit may have occurred when $B_{\Delta t} \in \Omega$. If a Brownian walker exits the domain, we approximate the exit position on the boundary $\partial \Omega$ by the intersection of line segment between $B_{0}$ and $B_{\Delta t}$ and the boundary $\partial \Omega$. We also approximate the exit time by linearly interpolating in time according to the approximated exit position in the line segment between $B_0$ and $B_{\Delta t}$. The approximation of the exit position and time is used in calculating the target value in Eq.~\eqref{eq:target_value} instead of the value of $u$ from the neural network. This target value is rich in the information of the exact solution since the exact value of function is given from the boundary condition. To enhance the information available on the boundary, we included an additional term in our loss function,
\begin{equation}\label{eq:loss_boundary}
\mathcal{L}^{\partial \Omega}(\theta) = \sum \limits_{k=1}^{S}\left(u(x_k;\theta)-h(x_k) \right)^2.
\end{equation}        

We train the neural network by minimizing the loss function,
\begin{equation}\label{eq:loss_total}
\mathcal{L}_n(\theta) := \mathcal{L}_n^{\Omega}(\theta) +\mathcal{L}^{\partial \Omega}(\theta),
\end{equation}
by sequentially updating the parameters with stochastic gradient descent,
\begin{equation}
\theta_n = \theta_{n-1}-\alpha \nabla_{\theta}\mathcal{L}_n(\theta_{n-1}). 
\end{equation}
The learning rate $\alpha$ may be adjusted on each step. In practice, this gradient descent step can be optimized to take previous steps into account, for example by using the ADAM optimization method~\cite{ADAM}.

\newpage
\renewcommand{\thempfootnote}{\fnsymbol{mpfootnote}}
\noindent\begin{minipage}{\textwidth}
\renewcommand\footnoterule{}
\begin{algorithm}[H]
  \caption{Derivative-Free Loss Method (DFLM): provides an estimate for the solution of the PDE~Eq.~\eqref{eq:quasilinear_elliptic} with boundary condition~Eq.~\eqref{eq:BCDirichlet} using a neural network and Brownian walkers. \label{algo:thealgo}}
  \begin{algorithmic}[1]
\Require{step-size $\Delta t$, learning rate $\alpha$, number of walkers $N$, number of Brownian samples per walker $M$, number of samples $S$, initial network parameters $\theta_0$ }
\Statex
\Function{DFLM}{$\Delta t, \alpha$, $N$, $M$, $S$, $\theta_0$}
\For{$1 \leq i \leq N$} 

\Let{$B_i$}{$Unif(\Omega)$}  \Comment{initialize walkers randomly}
\EndFor\vspace{3mm}
\For{$n=1,2,\ldots$} \Comment {each iteration}
    \For{$1 \leq i \leq N$, $1 \leq j \leq M$} 
    \Let{$B_{i,j}$}{$B_i + \sqrt{\Delta t} \mathcal{N}(0,I_d)$}
    \Comment{take $M$ Gaussian steps from each $B_i$\footnote{If a sampled position $B_i + \sqrt{\Delta t} \mathcal{N}(0,I_d)$ is outside of the domain $\Omega$, then $B_{ij}$ is projected onto the boundary $\partial \Omega$ as illustrated in Section~\ref{subsec:algorithm_explained}. }}
    \EndFor
    \For{$1 \leq i \leq N$}
    \Let{$y_i$}{$\frac{1}{M} \displaystyle\sum_{j=1}^M \left(u_\theta\cdot D(F,u_\theta)-R(G,u_{\theta})\right)\big|_{B_{i,j}}$ } 
   \Comment{estimate $y_i$ using Eq.~\eqref{eq:target_value}} 
    \EndFor
    
    \Let{$\mathcal{L}^{\Omega}_n(\theta)$}{$\frac{1}{N}\displaystyle\sum_{i=1}^N \frac{1}{2}\left(y_i - u_\theta(x_i) \right)^2$}
    \Comment{evaluate the loss function $\mathcal{L}^{\Omega}_n$ using Eq.~\eqref{eq:loss_inner}}\vspace{5mm}
    
    \For{$1 \leq k \leq S$}
    \Let{$x_k$}{$Unif(\partial \Omega)$} \Comment{sample on the boundary}
    \EndFor
      
	\Let {$\mathcal{L}^{\partial \Omega}(\theta)$}{$\displaystyle\sum_{k=1}^S(u_{\theta}(x_k)-h(x_k))^2$}
	\Comment{evaluate the loss function $\mathcal{L}^{\partial \Omega}$ using Eq.~\eqref{eq:loss_boundary}} \vspace{5mm}
	
	\Let {$\mathcal{L}_n(\theta)$}{$\mathcal{L}_n^{\Omega}(\theta)+\mathcal{L}^{\partial \Omega}(\theta)$}
    \Comment{evaluate the loss function $\mathcal{L}_{n}$ using Eq.~\eqref{eq:loss_total}}   \vspace{3mm}
    
    \Let{$\theta_{n+1}$}{$\theta_{n} - \alpha \frac{\partial}{\partial \theta}\mathcal{L}_n(\theta){\Big|}_{\theta=\theta_n}$ }
    \Comment{gradient descent update} \vspace{3mm}
	
	\For{$1 \leq i \leq N$}
	\Let {$B_i$}{$B_i+\sqrt{\Delta t}\mathcal{N}(0,I_d)$}
	\Comment{move the Brownian walkers\footnote{If a sampled position $B_i + \sqrt{\Delta t} \mathcal{N}(0,I_d)$ is outside of the domain $\Omega$, then $B_i$ is reinitialized to a uniformly random location in the domain $\Omega$ (i.e., $B_i \leftarrow Unif(\Omega)$).}}
    \EndFor
    \EndFor
    
    \EndFunction
    
  \end{algorithmic}
\end{algorithm}
\end{minipage}
\newpage
\section{Examples}\label{sec:Examples}

In this section we use the DFLM to compute numerical solutions to some example quasilinear elliptic differential equation in the form of Eq.~\eqref{eq:quasilinear_elliptic}. We measure the accuracy of numerical solutions by the relative $\mathcal{L}_2$-error, $\frac{\|u_{\theta}-u\|_{2,\Omega}}{\|u\|_{2,\Omega}}$ if the analytic solution is known (examples in Section~\ref{subsec:Laplace} and \ref{subsec:interface}), and we compute the relative $\mathcal{L}_2$-difference between the numerical solution from our method and that from the finite element method (FEM) if the analytic solution is unknown (the example in Section~\ref{subsec:taxis}).

We implement our algorithm using Tensorflow-GPU~\cite{TENSORFLOW}, which are open source libraries for deep learning. In particular, Tensorflow is capable of automatic differentiation (AD) of functions specified by a computer program and, for instance, it can calculate the gradients of a parameterized function $u(x;\theta)$, a neural network in this work, with respect to both $x$ and $\theta$.

We emphasize that our algorithm does not directly compute the derivatives of the neural network, $\frac{\partial u_{\theta}}{\partial x_i}$ or $\frac{\partial^2 u_{\theta}}{\partial x_i \partial x_j}$, and we only use the automatic differentiation in computing the gradients of the loss function, $\nabla_{\theta}\mathcal{L}_n(\theta)$. In Section~\ref{subsec:Laplace}, we present some empirical results demonstrating how this is a benefit when solving Laplace's equation. The example in Section~\ref{subsec:interface} is an elliptic interface problem which we use to demonstrate how our algorithm deals with a discontinuous solution and discuss the benefits of our use of the Cameron-Martin-Girsanov theorem. In Section~\ref{subsec:taxis}, we consider a practical example of a differential equation that is a model of chemotaxis.

\subsection{Laplace's equation}\label{subsec:Laplace}
As a representative example of a problem in elliptic PDEs, we apply our method to solve Laplace's equation with a Dirichlet boundary condition,
\begin{equation}\label{eq:Laplace_eq}
\begin{split}
\Delta u &= 0 \textrm{~in~} \Omega, \\
u &= h \textrm{~on~}  \partial \Omega. \\
\end{split}
\end{equation}
The domain $\Omega \in \mathbb{R}^2$ is a circular sector $\Omega=\{(r,\theta) : 0\leq r \leq 1,~0\leq \theta \leq \frac{\pi}{6}\}$ and $h$ is the given boundary values. We choose this boundary data by evaluating a harmonic function in $\Omega$ on $\partial \Omega$. This provides us with an exact solution to the problem. We use two different harmonic functions $u_1(x_1,x_2)=x_1^2-x_2^2-\frac{1}{4}x_1x_2$ and $u_2(r,\theta)=r^{\frac{2}{3}}\sin\left(\frac{2}{3}\theta \right)$. Note that $u_1$ is smooth on $\Omega \cup \partial \Omega$, but the derivatives of $u_2$, $\partial_ru_2$ and $\partial^2_ru_2$, have a singularity at the origin. The elliptic problem with $u_2$ as the solution is known as a \textit{corner singularity} problem \cite{CORNERSINGULARITY}, which is a benchmark for testing new numerical algorithms.

The Bellman equation corresponding to Eq.~\eqref{eq:Laplace_eq} is, for any $t_0 < t$,
\begin{equation}\label{eq:Bellman_eq1}
u(B_{t_0};\theta) = \e\left[u(B_{t};\theta) \right].
\end{equation} 
Our method is to find $\theta$ that minimizes the residual of Eq.~\eqref{eq:Bellman_eq1}.

For these test problems, we compare our method with another method~\cite{DGM,PINN,raissi2018forward,raissi2018deep} that uses a deep neural network that is explicitly differentiated with respect to its input neurons. We refer to this method as the derivative-based loss method (DBLM), which essentially finds $\theta$ that minimizes the residual of
\begin{equation}
\partial^2_{x_1}u(x;\theta) + \partial^2_{x_2}u(x;\theta) = 0, \hspace{3mm}\forall x=(x_1,x_2) \in \Omega.
\end{equation}
Note that $\partial^{2}_{x_i}u(x;\theta)$, $i=1,2$ are derivatives of the neural network which can be computed using automatic differentiation, i.e backpropagation.
The parameters $\theta$ are founded by stochastic gradient method using the loss function at each iteration, 
\begin{equation}
\mathcal{L}(\theta) = \sum\limits_{i=1}^{N}(\partial^2_{x_1}u(x_i;\theta) + \partial^2_{x_2}u(x_i;\theta))^2 + \sum \limits_{k=1}^{S}\left(u(z_k;\theta_n)-h(z_k) \right)^2
\end{equation}
where $x_i$, $i=1,2,\cdots,N$ are mini-batch samples in $\Omega$ and $z_k$, $k=1,2,\cdots, S$ are mini-batch samples on $\partial \Omega$. 
The DFLM uses the positions of $N$ independent Brownian motions at each discrete time $n\Delta t$ as minibatch samples in $\Omega$. In the DBLM, uniform minibatch sampling on $\Omega$ at each iteration is used since it yields faster training speed than using Brownian walker sampling.

\begin{figure}[!]
\centering
\includegraphics[width=1.0\textwidth]{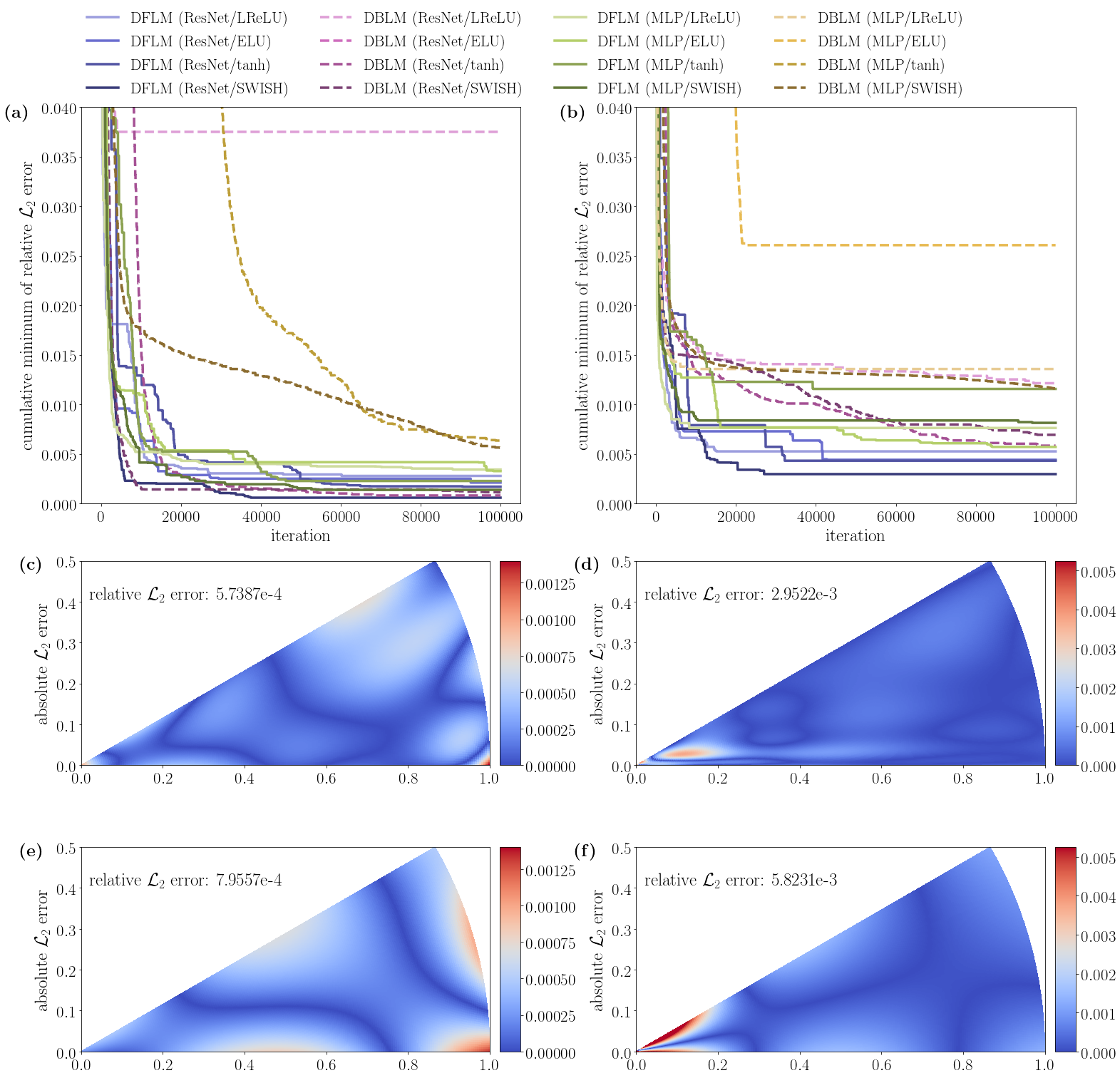}
\caption{Numerical solutions to Dirichlet problem for Laplace's equation Eq.~\eqref{eq:Laplace_eq}. The exact solutions are $u_1(x_1,x_2)=x_1^2-x_2^2-\frac{1}{4}x_1x_2$ (the first column, (a), (c), and (e)) and $u_2(r,\theta)=r^{\frac{2}{3}}\sin\left(\frac{2}{3}\theta \right)$ (the second column, (b), (d), and (f)). The first row, (a) and (b),  presents the cumulative minimum of relative $\mathcal{L}_2$-errors during $10^5$ iterations in training of 8 different neural networks using the derivative-free loss method (solid lines) and the derivative-based loss method (dashed lines). The second row, (c) and (d), shows the pointwise error of the numerical solutions using the derivative-free loss method with the highest accuracy among the 8 neural networks, the ResNet with SWISH activation. The third row, (e) and (f), shows the pointwise error of the numerical solutions using the derivative-based loss method with the highest accuracy among 8 neural networks, the ResNet with tanh activation.
}
\label{fig:Laplace}
\end{figure}

In order to compare the two methods, we use the multilayer perceptron and a variant of ResNet (defined in Section~\ref{subsec:NeuralNetwork}) with 4 different activation functions, LReLU,  ELU, $\tanh$, and SWISH~\cite{SWISH}. All hyperparameters are identical between the two methods. The MLP has the dimension $[2,20,20,20,20,1]$ and the ResNet is constructed as the stacks of 3 identical residual blocks with the dimension $[20,20,20]$. The hyperparameters, $N=1500$ (the number of Brownian walkers in the DFLM and mini-batch samples in $\Omega$ for the DBLM), $S=300$ (the number of mini-batch samples on $\partial \Omega$ for both methods), $M=200$ (the number of samples for the estimation of the targets in the DFLM) and $\Delta t =$5.0e-4 (the discrete timestep of Brownian motion in the DFLM) are chosen, and the ADAM optimization method with an exponentially decaying learning rate is used to train the neural networks. The relative $\mathcal{L}_2$-errors of the numerical solutions after $10^5$ iterations is presented in Table~\ref{tab:Laplace}. The table also shows the error when the network is trained directly using samples of the exact solution. More detailed results are presented in Fig.~\ref{fig:Laplace}.

The ResNet is more effective than MLP for both methods, and when using the DFLM, the ResNet with SWISH activation approximates the solutions $u_1$ and $u_2$ with the highest accuracy. The convergence of the DFLM is relatively insensitive to the choice of activation function as can be seen in Fig.~\ref{fig:Laplace} (a) and (b)). Also, in this example, the training of our DFLM converges in less (wall-)time than the DBLM.

\begin{center}
\begin{table}
\centering
\begin{tabular}{ccccc|ccc}
& & \multicolumn{3}{c @{}}{$u_1(x_1,x_2)=x_1^2-x_2^2-\frac{1}{4}x_1x_2$}&\multicolumn{3}{c @{}}{$u_2(r,\theta)=r^{\frac{2}{3}}\sin\left(\frac{2}{3}\theta \right)$}\\
\cmidrule(l){3-5} \cmidrule(l){6-8} 
Network &\text{Activation} & DFLM & DBLM & Direct & DFLM & DBLM & Direct\\
\hline
\multirow{4}{*}{MLP}
&LReLU & 3.41e-3 & 4.35e-2 & 7.81e-4 & 7.61e-3 & 1.36e-2 & 1.06e-3\\
&ELU   & 3.26e-3 & 6.78e-2 & 8.33e-4 & 5.70e-3 & 2.61e-2 & 1.25e-3 \\
&tanh  & 2.26e-3 & 6.25e-3 & 6.94e-4 & 1.12e-2 & 5.15e-2 & 1.09e-3 \\
&SWISH & 1.36e-3 & 5.62e-3 & 1.77e-4 & 8.13e-3 & 1.12e-2 & 4.78e-4 \\
\hline
\multirow{4}{*}{ResNet}
&LReLU & 2.77e-3 & 3.75e-2 & 3.34e-4& 5.25e-3 & 1.21e-2 & 8.22e-4 \\
&ELU   & 2.10e-3 & 1.07e-1 & 5.19e-4 &4.40e-3 & 4.16e-2 & 8.13e-4 \\
&tanh  & 1.68e-3 & \bf{7.96e-4} & 2.69e-4 & 4.30e-3 & \bf{5.82e-3} & 9.53e-4 \\
&SWISH & \bf{5.74e-4} & 1.11e-3 & 1.69e-4 & \bf{2.95e-3} & 6.92e-3 & 2.23e-4 \\
\bottomrule
\end{tabular}
\caption{Relative $\mathcal{L}_2$-errors of the approximated solutions the DFLM, the DBLM which includes the differentiation of a neural network, and direct approximation of the exact solution, within $10^5$ iterations. }
\label{tab:Laplace}
\end{table}
\end{center}

\subsection{An interface problem} \label{subsec:interface}
Interface problems arise when modeling diverse physical and biological phenomena such as electrostatics in composite materials, multiphase flow in fluid dynamics, heat conduction, and the electrical activity of biological cells. In this section, we solve the elliptic equation governing the electric potential on a domain with a single interface. Our method requires a small modification to account of the interface conditions. Although our example has only a single simple interface, our method is easily applied to problems with multiple, possible intricate, interfaces.

Let $\Omega$ be the bounded domain in $\mathbb{R}^2$, which consists of two subdomains $\Omega_0=\{(r,\theta):0\leq r < 1\}$ and $\Omega_1=\{(r,\theta):1<r<2\}$ with the interface $\Gamma=\{(r,\theta):r=1\}$. The conductivity $\sigma$ on each $\Omega_i$ has constant value $\sigma_i$ for $i=0,1$ and the potential difference $\Delta V$ is given across the interface $\Gamma$. The electric potential function $u(x)$ satisfies
\begin{align}
\nabla \cdot\left( \sigma \nabla u \right) &= g \textrm{~in~} \Omega_0 \cup \Omega_1, \label{eq:interface_eq}\\
u &= h \textrm{~on~} \partial \Omega,\label{eq:interface_bdry} \\
[u] &= \Delta V \textrm{~on~} \Gamma,\label{eq:interface_jump}\\
\left[\sigma \frac{\partial u}{\partial n}\right] &= 0 \textrm{~on~} \Gamma. \label{eq:interface_continuity}    
\end{align}
The function $g$ in Eq.~\eqref{eq:interface_eq} is the given source distribution over $\Omega$ and the function $h$ in Eq.~\eqref{eq:interface_bdry} is the potential on the boundary $\partial \Omega$. Equation~\eqref{eq:interface_jump} is the jump condition resulting from a voltage difference across $\Gamma$. Equation.~\eqref{eq:interface_continuity} is the flux continuity condition across the interface $\Gamma$. Here $[q]$ is defined as $[q(x_0)]=q^{+}(x_0)-q^{-}(x_0)$ where $q^{+}(x_0)$ is the limit of $q(x)$ as $x \rightarrow x_0$ from $\Omega_1$ and $q^{-}$ is from $\Omega_0$. 

To appreciate the idea of the modification we make to our method to satisfy the conditions Eq.~\eqref{eq:interface_jump} and Eq.~\eqref{eq:interface_continuity}, consider the Bellman equation corresponding to  Eq.~\eqref{eq:interface_eq} without applying the Cameron-Martin-Girsanov theorem, as discussed in Eq.~\eqref{eq:linear_Martingale_process} and \eqref{eq:linear_Bellman_general},  
\begin{equation}\label{eq:interface_bellman_no_cm}
u(X_{0}) = \e\left[u(X_{\Delta t})-\int_{0}^{\Delta t}g(X_s) \text{d} s \right].
\end{equation}
The Brownian walkers take steps from $X_0$ and $X_{\Delta t}$ according to
\begin{equation}\label{eq:interface_stochastic_process_no_cm}
X_{\Delta t} = X_0+\nabla\sigma(X_0) \Delta t + \sqrt{2\sigma(X_0)\Delta t}\mathcal{N}(0,I_2),
\end{equation}
for a time step $\Delta t$. For $X_0$ in $\Omega_0$ ($\Omega_1$) near the interface $\Gamma$, if $X_{\Delta t}$ is in $\Omega_1$ ($\Omega_0$), the value $u(X_{\Delta t})$ is approximately equal to $u(X_0)+\Delta V$ ($u(X_0)-\Delta V$) by the jump condition Eq.~\eqref{eq:interface_jump}. Thus, we modify the Bellman equation Eq.~\eqref{eq:interface_bellman_no_cm} to   
\begin{equation}\label{eq:interface_bellman_no_cm_jump}
u(X_{0}) = \e\left[\bigg\{u(X_{\Delta t})- \Delta V \left(\one_{\Omega_0}(X_0) - \one_{\Omega_0}(X_{\Delta t})\right)\bigg\}-\int_{0}^{\Delta t}g(X_s) \text{d} s \right].
\end{equation}  

Next, in order to address the continuity condition Eq.~\eqref{eq:interface_continuity}, let's assume $X_0$ in $\Omega_0$ near the interface $\Gamma$. Since $\sigma$ is constant on $\Omega_0$ as $\sigma_0$, 
\begin{equation}\label{eq:interface_process_discontinuous_sigma}
X_{\Delta t}=X_0+\sqrt{2\sigma_0\Delta t}\mathcal{N}(0,I_2),
\end{equation}
in which the distribution of $X_{\Delta t}$ is symmetric with respect to $X_0$ and only depends on the conductivity $\sigma_0$ of $\Omega_0$ in which $X_0$ lies. However, this is not the case for the continuous stochastic process. During the continuous time $\Delta t$, once a Brownian walker crosses the interface to $\Omega_1$, it moves according to the conductivity $\Omega_1$. For instance, if $\sigma_1 > \sigma_0$, the walker moves faster in $\Omega_1$ than in $\Omega_0$ and, approximately, the distribution of $X_{\Delta t}$ has a longer tail in $\Omega_1$ than in $\Omega_0$. Moreover, the distribution of $X_{\Delta t}$ depends on the distance between $X_0$ and the interface $\Gamma$ but the distribution from the Eq.~\eqref{eq:interface_process_discontinuous_sigma} only depends on whether $X_0$ is $\Omega_0$ or in $\Omega_1$. To capture this property of the continuous stochastic process, we regularize the discontinuity of the conductivity function on the interface $\Gamma$. We define a regularized conductivity function $\sigma^{\epsilon}$ that converges to $\sigma$ as $\epsilon\rightarrow 0$ using the sigmoid function,
\begin{equation}
\sigma^{\epsilon}(\mathbf{x}) = \frac{\sigma_1-\sigma_0}{1+\exp(-(\|\mathbf{x}\|-1)/\epsilon)}+\sigma_0, \hspace{2mm} \epsilon > 0.
\end{equation}
We therefore move the walkers according to 
\begin{equation}\label{eq:interface_stochastic_process_regularized_no_cm}
X_{\Delta t} = X_0+\nabla\sigma^{\epsilon}(X_0) \Delta t + \sqrt{2\sigma^{\epsilon}(X_0)\Delta t}\mathcal{N}(0,I_2).
\end{equation}

There is a strong drift $\nabla \sigma^{\epsilon}$ near the interface toward the region with higher conductivity and the drift is increasing closer to the interface. The solution $u^{\epsilon}$ of Eq.~\eqref{eq:interface_eq}, Eq.~\eqref{eq:interface_bdry}, and Eq.~\eqref{eq:interface_jump} with the regularized conductivity $\sigma^{\epsilon}$ automatically satisfies the continuity condition $\left[\sigma^{\epsilon}\frac{\partial u^{\epsilon}}{\partial n} \right]$ and $u^{\epsilon}$ is an approximation of the exact solution $u$.

The parameter $\epsilon$ should be selected so that, with high probability, the walkers sample the region where the drift $\nabla \sigma^{\epsilon}(x)$ is large. If $\epsilon$ is too small relative to the typical distance travelled by the walkers, $\sqrt{2\sigma^{\epsilon}(X_0)\Delta t}$, the information about the regularized function near the interface goes unnoticed. 

This example illustrates an advantage of using the Cameron-Martin-Girsanov theorem particularly well. Without writing the problem in terms of ordinary Brownian motion, the walkers move as Eq.~\eqref{eq:interface_stochastic_process_regularized_no_cm} and their movement depends on the conductivity of their current positions. This dependence is problematic. If the conductivity is too small, the walkers move very slowly on that subdomain and no longer efficiently explore it. Moreover, if one conductivity is very large relative to the other conductivity, the walkers are effectively `absorbed' into the subdomain with small conductivity causing an imbalance of sampling between the two subdomains. Furthermore, the walkers near the interface tend to move toward the direction of drift pushing them away from the interface, which slows the neural network from learning the information about the interface. The Cameron-Martin-Girsanov theorem simplifies the walkers' motion to that of ordinary Brownian motion which is able to explore the domain without interference from the PDE. All of the information from PDE is encoded in the discounts and rewards in the Bellman equation. This is akin to the method of importance sampling in Monte Carlo methods.

In summary, we apply our method to solve the interface problem with the following modified Bellman equation, which includes a term for the jump $\Delta V$,
\begin{multline}
u(B_0)= \e \bigg[ \bigg\{ u(B_{\Delta t})- \Delta V \left(\one_{\Omega_0}(B_0) - \one_{\Omega_0}(B_{\Delta t})\right) \bigg\}
\\ \cdot\exp\left(\int_0^{\Delta t}\frac{\nabla \sigma^{\epsilon}}{2 \sigma^{\epsilon}}\cdot dB_s - \half \int^{ t}_0 \left\|\frac{\nabla \sigma^{\epsilon}}{2 \sigma^{\epsilon}} \right\|^2 \text{d} s \right) - \int^{t}_0 \frac{g}{2\sigma^{\epsilon}}\text{d} s \bigg].
\end{multline}

Considering that the solution is not continuous or differentiable across the interface, we construct the parameterized approximation of the solution separately on each subdomain $\Omega_0$ and $\Omega_1$. This is implemented simply in a neural network by adding a categorical variable of subdomains in the input layer. We encode the categorical variable by a one-hot vector. The input layer is 4 dimensional vector in which the first 2 coordinates represent $x$, and last 2 coordinates are the one-hot vector. In particular, $x=(x_1, x_2)$ in $\Omega_0$ is represented by $(x_1, x_2, 1, 0)$ and $x=(x_1, x_2)$ in $\Omega_1$ by $(x_1, x_2, 0, 1)$. This is efficient in that a single neural network can represent multiple functions on multiple subdomains. Since a neural network could approximate discontinuous or non-differentiable functions, since it is a universal approximator, this variation of input layer is not crucial in our methodology. However, as more information about the solution is reflected in the neural network, it is more accurate and trains faster.

\begin{figure}[!] 
\centering
\includegraphics[width=1.0\textwidth]{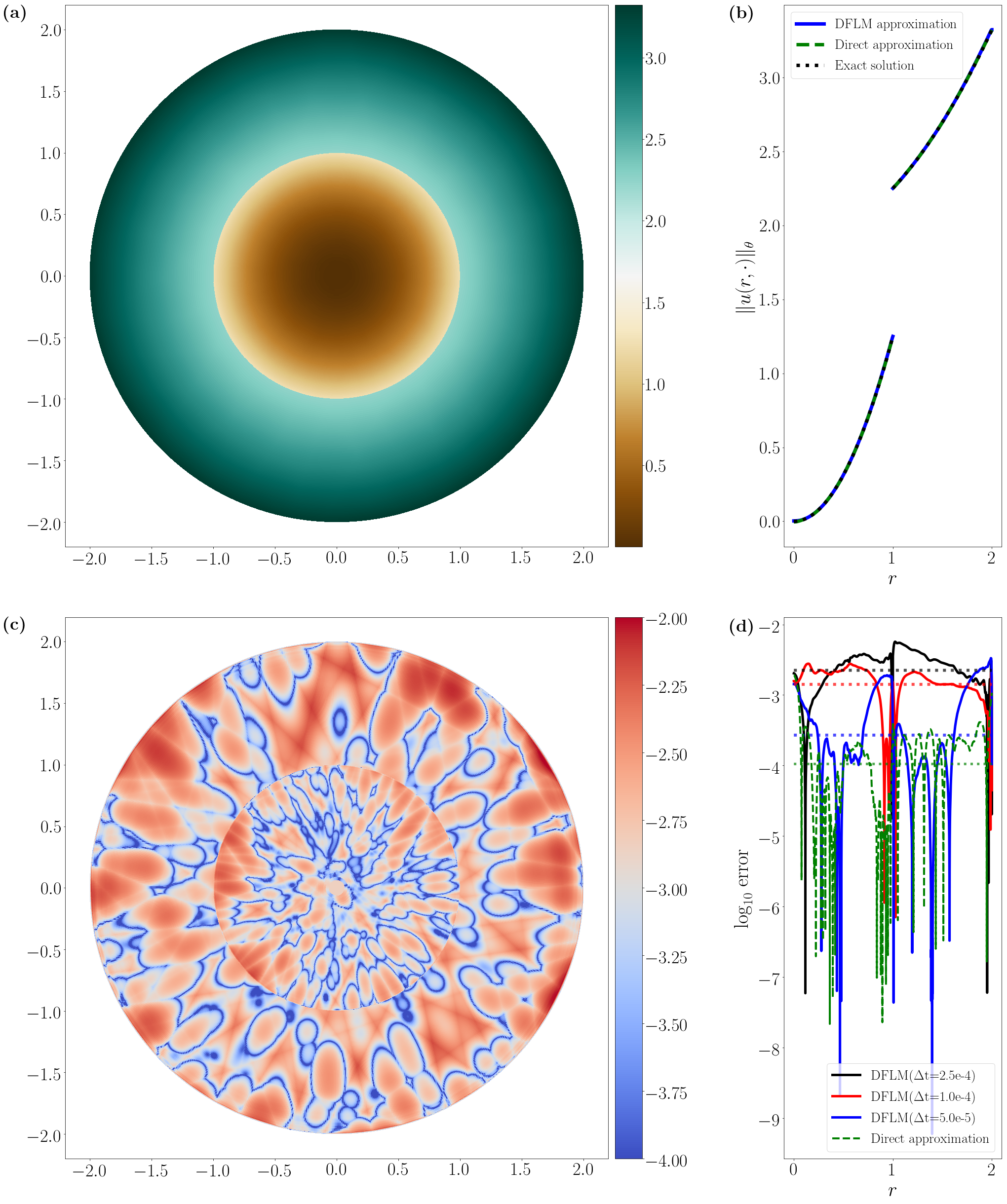}
\caption{The numerical solutions of the interface problem Eqs.~\eqref{eq:interface_eq}~-~\eqref{eq:interface_continuity}, which has the exact solution Eq.~\eqref{eq:interface_exact}. We trained the ResNet described in the main text with 3 different time steps $\Delta t =$2.5e-4, 1.0e-4, and 5.0e-5 and also directly trained to the exact solution. As the time step is reduced, the neural network approximates the solution with higher accuracy and it has relative $\mathcal{L}_2$-error 2.6975e-4 with $\Delta t =$ 5.0e-4. The first row presents (a) the numerical solution (with $\Delta t=$5.0e-5)  on the domain, and (b) the circular averages $\|u(r,\cdot)\|_{\theta}$ of the numerical solution, the neural network approximation directly trained to the exact solution, and the exact solution. The second row presents (c) the relative error of the numerical solution (with $\Delta t=$5.0e-5) on a $\log_{10}$ scale and (d) the relative errors of the circular averages of the 3 numerical solutions and the direct approximation on a $\log_{10}$ scale.}
\label{fig:interface}
\end{figure}

We test our method with $g=1$, $h=1+\frac{1}{4\sigma_0}+\frac{3}{4\sigma_1}$, $\Delta V =1$, $\sigma_0=0.2$, $\sigma_1=0.7$, for which the analytic solution is 
\begin{equation}\label{eq:interface_exact}
u(r)=
\begin{cases}
\frac{1}{4\sigma_0}r^2, & r < 1, \\
\frac{1}{4\sigma_1}r^2+\left(1-\frac{1}{4\sigma_1}+\frac{1}{4\sigma_0} \right), & 1 < r < 2.
\end{cases}
\end{equation}
Note that we do not use the knowledge about the exact function being a radial function and the Cartesian coordinates are used in the input layer. Based on our experiment, the variant of ResNet outperform the MLP and the structure of ResNet with LReLU activation in the first layer activation ($\sigma^{\text{in}}$ defined in Eq.~\eqref{eq:resnet_structure}) and SWISH ($\beta=1$) activation in the residual blocks is suitable for this example. In particular, we suspect that the choice of LReLU activation in the first layer is effective for the neural network to represent the multiple functions based on the categorical input variable. The simulation results are presented in Fig.~\ref{fig:interface}.  The ResNet described above with 3 identical residual blocks with dimension $[60,60,60,60]$ is trained with $N=2000$ (the number of Brownian walkers), $S=200$ (the number of mini-batch samples on the boundary $\partial \Omega$), $M=200$ (the number of samples for the estimation of the target) by the ADAM optimization method. We measured the accuracy of the neural network approximation using a radial average with $T=10^{4}$ angular samples, $u_{\theta}(r):=\frac{1}{T}\sum \limits_{k=1}^{T}u\left(r\cos\frac{2\pi}{T}k, r\sin\frac{2\pi}{T}k;\theta\right)$, $r \in [0,2]$. We trained the neural network with different time steps $\Delta t$ of the Brownian motion. The neural network approximates the radial dependency of the solution, the jump condition, and the flux continuity condition on the interface, in total, with relative $\mathcal{L}_2$-error 2.6975e-4. The details are presented in Fig.~\ref{fig:interface}. 
\\

\subsection{A steady-state population model with taxis}\label{subsec:taxis}
In this section, we consider a quasi-linear elliptic equation used in the modeling of \textit{taxis}, which is the movement of living systems in response to external stimulus. For example, \textit{phototaxis} refers to the movement of motile organism toward or away from the source of light, and \textit{chemotaxis} is the migration of motile cell or organism in a direction affected by the gradient of diffusible substance. The phenomenological equation that describes the steady-state population density $u(x)$ with taxis stimulus $c(x)$ is 
\begin{equation}\label{eq:taxis_equation}
\nabla(D\nabla u +G(u, c)) + H(u) =0,
\end{equation}
in which $D$ is the diffusion coefficient, $G(u,c)$ is the taxis flux, and $H(u)$ is the kinetics of the population. In particular, we choose the  taxis flux $G(u,c)$ as $G(u,c)=-\chi u\nabla c$, the diffusive flux of stimulus concentration with a coefficient proportional to the population. This choice of taxis flux is motivated from the model of chemotaxis, known as the Keller-Segel model \cite{KSMODEL,KSMODEL2}. For given $c(x)$ and $H(x)$, the Bellman equation corresponding to Eq.~\eqref{eq:taxis_equation} is 
\begin{multline}\label{eq:taxis_Bellman}
u(B_0) = \e\bigg[u(B_{\Delta t})\exp\left(-\int_0^{\Delta t}\frac{\chi}{2D}\nabla c(B_s)\cdot \text{d} B_s-\frac{1}{2}\int_{0}^{\Delta t}\left\|\frac{\chi}{2D}\nabla c(B_s) \right\|^2\text{d} s \right) \\
-\frac{1}{2D}\int^{\Delta t}_0\chi u(B_s) \Delta c(B_s) - H(u(B_s))\text{d} s \bigg].
\end{multline}

\begin{figure}[!]
\centering
\includegraphics[width=1.0\textwidth]{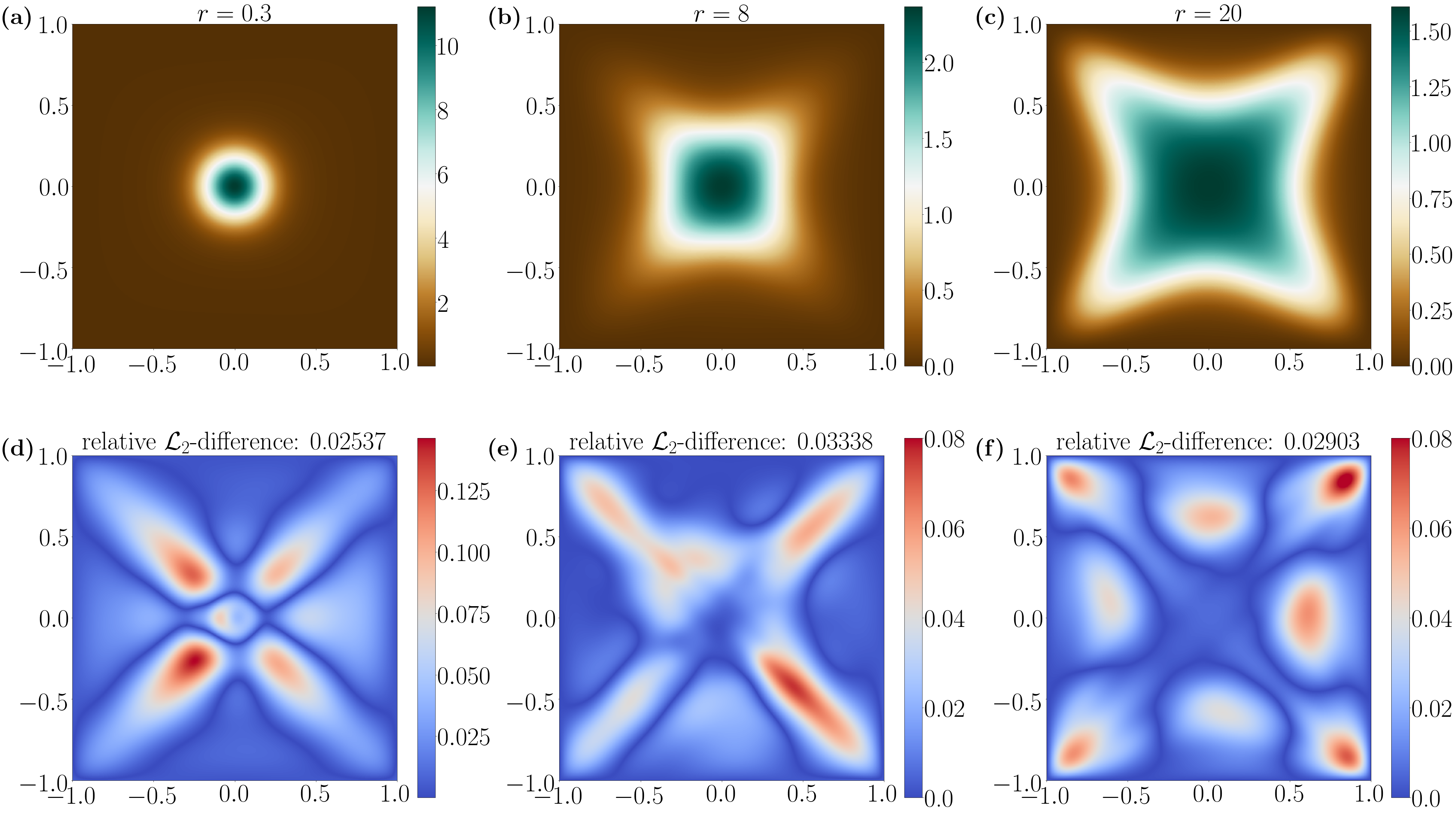}
\caption{Numerical solutions of the population model equation Eq.~\eqref{eq:taxis_equation} with the taxis stimulus $c(x)=\frac{1}{2}\sin\left(\frac{\pi}{2}(x_1+1) \right)\sin\left(\frac{\pi}{2}(x_2+1) \right)$ and three logistic growth rates $r=0.3$ in (a), $r=8$ in (b), $r=20$ in (c). The remaining parameters are $\chi=5$, $r_0=0.5$, $D=0.1$. The population aggregates toward the peak of the stimulus and the peak of the population reduces as the logistic growth rate increases. The dependence of the peak value on the logistic growth rate $r$ is shown in the Fig.~\ref{fig:kellerSegel_2} (a). The solutions are compared with those from the finite element method. The absolute pointwise difference is shown in (d) for $r=0.3$, (e) for $r=8$, and (f) for $r=20$. The numerical solutions are computed using the ResNet with stacks of 3 identical residual blocks with dimensions $[40,40,40,40]$, an ELU activation function, $N=2000$ Brownian walkers, $S=200$ boundary samples, $M=200$ samples for computing the target, and $\Delta t =$5.0e-4.}
\label{fig:kellerSegel}
\end{figure}

We solved the equation in the domain $\Omega = [-1,1]^2$ with the taxis stimulus function, $c(x_1,x_2)=\frac{1}{2}\sin\left(\frac{\pi}{2}(x_1+1) \right)\sin\left(\frac{\pi}{2}(x_2+1) \right)$, in which the stimulus has the peak at $(x_1,x_2)=(0,0)$. We choose quadratic population kinetics, $H(u)=ru(1-u)+r_0$, in which $r$ is the logistic growth rate and $r_0$ is the constant growth rate. We set the population to be zero on the boundary of the domain (i.e., $u=0$ on $\partial \Omega$). The simulation results with different logistic growth rates are presented in Fig. \ref{fig:kellerSegel}. 
The results show that the smaller the logistic growth rate is, the more the population aggregates near the peak(s) of taxis stimulus functions. The numerical results are compared to the results from the finite element method using the MATLAB PDE Toolbox~\cite{MATLABPDE} by the relative $\mathcal{L}_2$-difference measurement.

Since taxis models typically display diverse phenomenon as their parameters are varied, we present a modification of our method that finds a parameter-dependent family of numerical solutions. In particular for this example, we compute the solution $u(x;r)$, which denotes the solution of Eq.~\eqref{eq:taxis_equation} for the logistic growth rate $r$ from the kinetics function $H(u)$, within a bounded range $[r_{\min}, r_{\max}]$. We regard the parameter $r$ as a new input variable and construct a neural network estimation $u(x;r;\theta)$, in which the input layer of the neural network is $(x,r)$. We train this neural network to minimize the residual of the Bellman equation Eq.~\eqref{eq:taxis_Bellman}. In each iteration of our algorithm, each Brownian walker is assigned a value of $r$ in $[r_{\min}, r_{\max}]$ and contributes a term to the loss function corresponding to the residual of Eq.~\eqref{eq:taxis_Bellman}. There are many options for assigning the value of $r$ to each Brownian walker at each iteration. We choose to treat the value of each walker's $r$ as a one dimensional Brownian motion in the domain $[r_{\min}, r_{\max}]$ to be consistent with the spirit of our algorithm.
Each $r$ is randomly initialized in the domain and moves as Brownian motion at each iteration step, that is, $r \leftarrow r + \sigma_{\textrm{r}}\sqrt{\Delta t}\mathcal{N}(0,1)$. When a walker's value of $r$ crosses the boundary ($r_{\min}$ or $r_{\max}$), it is reinitialized to a uniformly random location in the parameter domain. The value of $\sigma_{\textrm{r}}$ should be a small, but not very small, fraction of the size of the parameter domain. We use $\sigma_{\textrm{r}}=1$.

\begin{figure}[!]
\centering
\includegraphics[width=1.0\textwidth]{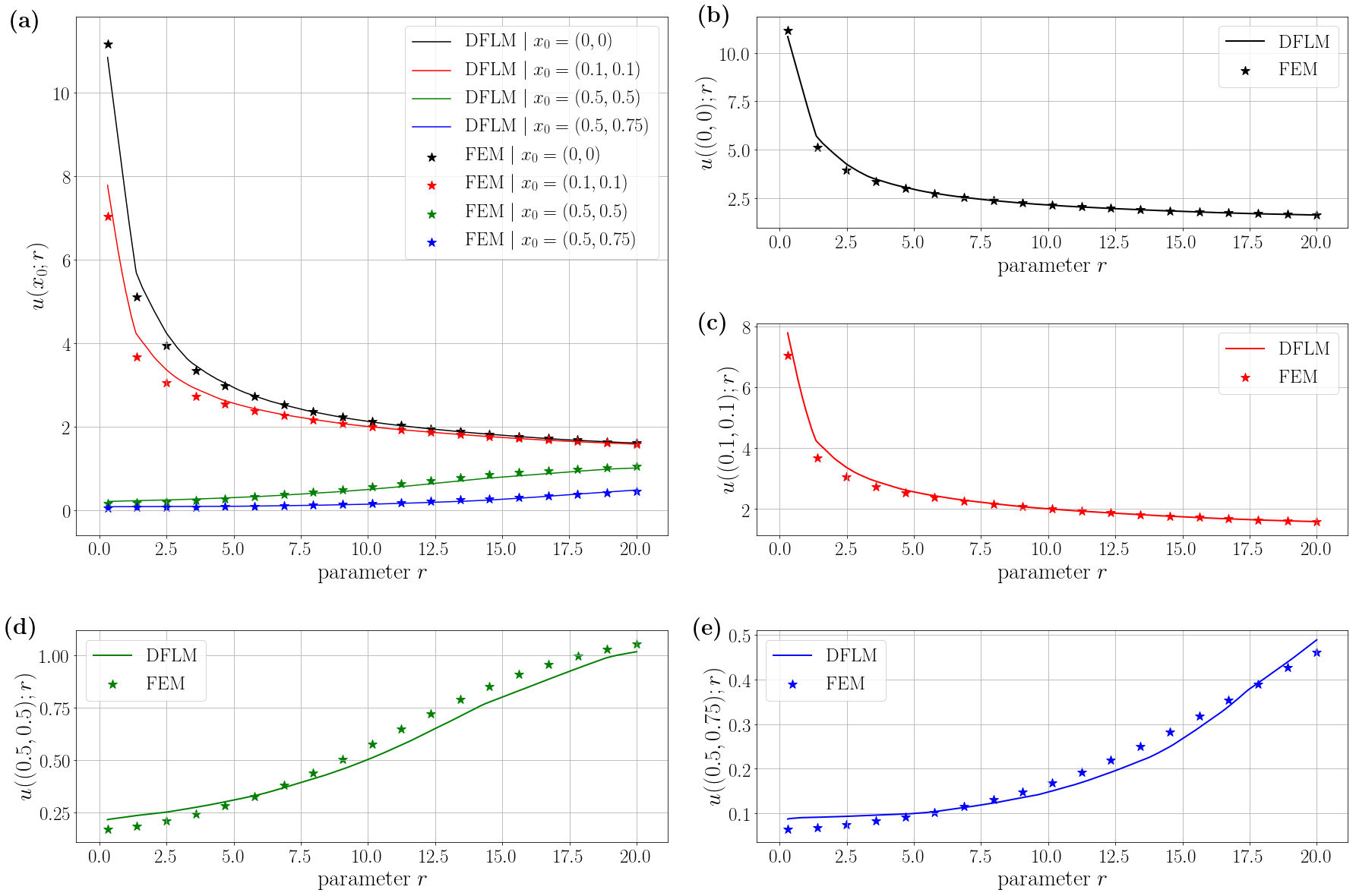}
\caption{The value of the solution at the selected points, as a function of the parameter $r$, the logistic growth rate, for (a) all of the selected points and (b)-(e) at each of the four selected points. The DFLM values (solid lines in (a)-(e)) are evaluated from the estimated family of solutions, $u(x;r;\theta)$ for $r\in[0.3,20]$. The FEM values (markers in (a)-(e)) are evaluated by solving the equation with each parameter $r$ separately. The family of solutions are estimated using the ResNet with stacks of 4 identical residual blocks with dimensions $[40,40,40,40]$ and the LReLU ($\alpha=0.1$) activation function. We used $N=2000$ Brownian walkers, $S=200$ boundary samples, $M=200$ samples for computing the target, and $\Delta t =$5.0e-4 to train the network.}
\label{fig:kellerSegel_2}
\end{figure}

Figure~\ref{fig:kellerSegel_2} shows the dependence of the solution of Eq.~\eqref{eq:taxis_equation} on the logistic growth rate $r$ for $r \in [0.3,20]$ using the DFLM and the FEM. We see that the two methods give similar values, but are unable to attribute the difference primarily to errors in either method. The computation time was significantly shorter for the DFLM compared to the FEM.

The neural network provides a quickly-evaluated interpolant of the solution's parameter dependence, which is useful for a variety of applications. In particular, it could be used to estimate the parameter from observed data by solving a nonlinear optimization problem (e.g. minimizing $\mathcal{L}(r)=\sum\limits_{k=1}^{K}|u(x_k;r)-y_k|^2$, in which $(x_k,y_k), k=1,2,\cdots, K$, are the observed data) without having to solve the PDE each time the loss is evaluated.

\section{Conclusions}

We constructed a numerical method based on sampling Brownian motion that trains neural networks to solve quasilinear elliptic PDEs without explicitly having to compute the derivatives of the neural network with respect to the input variables. There being numerous applications of quasilinear elliptic PDEs, we expect our robust, versatile, and efficient numerical method to be immensely useful. Like other neural network methods for PDEs, this method is grid-free and naturally parallelizable. In addition, by using Brownian motions, the method can handle prescribed jumps along an interface (as in Section~\ref{subsec:interface}) and avoids some issues related choosing sample points in the domain (as in the corner singularity example of Section~\ref{subsec:Laplace}). 

Our method was illustrated using two dimensional example problems, but it would be particularly effective for high dimensional problems. Explicitly computing the network derivatives and sampling in high dimensions is computationally expensive, while Brownian motions naturally explore space regardless of the dimension. Additionally, problems posed on infinite domains can be handled naturally within our method.

There are many future directions to explore with our method. One direction is to use other reinforcement learning methods to solve the Bellman equation, Eq.~\eqref{eq:linear_Bellman_general}. It is possible that a different method could be much more efficient than the simple $\mathcal{L}_2$ based loss function we use here. Other representations of the function beyond neural networks are compatible with our method primarily because it is not based on explicitly computing derivatives. By using Brownian motions, we have the option to handle boundary conditions by imposing conditions on the stochastic walkers, for example by reflecting them off the boundary for homogeneous Neumann boundary conditions, or by adapting the Bellman equation. Our method can also be improved and made more general by drawing on the rich theory of stochastic processes as we did with our use of the Cameron-Martin-Girsanov theorem. A generalization to systems of PDEs would be straightforward by having a neural network for each unknown or a single network with multiple outputs, one for each unknown.

Considerable work remains to adapt traditional ideas from the numerical solution of PDEs to our method. Extrapolation (in $\Delta t$), operator splitting, and domain decomposition would be especially beneficial to include.
A hybrid method, in which say finite elements or a boundary integral method were to be combined with our method, could be highly accurate and efficient. With regard to the use of neural networks, more complete studies of the training procedure and the network architecture best suited to solving partial differential equations are required.

Our method is readily applicable to quasilinear \textit{parabolic} PDEs by adding an input neuron for the time variable and adapting the Bellman equation to account for `exits' of the Brownian motion on the initial condition. In this case, the neural network would encode temporal information as it did for the parameter in the chemotaxis example. We hope to extend our approach to a wider class of partial differential equations by finding additional connections with stochastic processes.

Including a PDE problem's parameter-dependence in the neural network representation of the solution allows for efficient sensitivity analysis and regression-based data fitting. Our method is based on a Markov reward process, but could readily be extended to a Markov decision process which opens the possibility of solving numerical control and design problems.

\section*{Acknowledgments}
We acknowledge the support of the Natural Sciences and Engineering Research Council of Canada (NSERC): RGPIN-2019-06946 for A.R.S. and PDF-502287-2017 for M.N., as well as a University of Toronto Connaught New Researcher Award. We appreciate the input from the students participating in the 2019 Fields Undergraduate Summer Research Program: Julia Costacurta, Cameron Martin, and Hongyuan Zhang.

\bibliographystyle{elsarticle-num}
\bibliography{references}

\end{document}